\UseRawInputEncoding
\documentclass{article}
\pdfoutput=1%
\usepackage{microtype}
\usepackage{graphicx}
\usepackage{subcaption}
\usepackage{booktabs} % for professional tables
\usepackage{xcolor}
\usepackage{bbm}
\usepackage{makecell}
\usepackage{import}
\usepackage{float}

\usepackage{hyperref}

\usepackage[accepted]{icml2023}

\usepackage{amsmath}
\usepackage{amssymb}
\usepackage{mathtools}
\usepackage{amsthm}

\newcommand{\Ours}{GoBI}
\newcommand{\OursName}{Go Beyond Imagination}

\usepackage[capitalize,noabbrev]{cleveref}

\theoremstyle{plain}

\theoremstyle{definition}

\theoremstyle{remark}

\usepackage[textsize=tiny]{todonotes}

\icmltitlerunning{Go Beyond Imagination: Maximizing Episodic Reachability with World Models}

\begin{document}

\twocolumn[
\icmltitle{Go Beyond Imagination: Maximizing Episodic Reachability with World Models}



\icmlsetsymbol{equal}{*}

\begin{icmlauthorlist}
\icmlauthor{Yao Fu}{yyy}
\icmlauthor{Run Peng}{yyy}
\icmlauthor{Honglak Lee}{yyy,comp}
\end{icmlauthorlist}

\icmlaffiliation{yyy}{University of Michigan}
\icmlaffiliation{comp}{LG AI}

\icmlcorrespondingauthor{Yao Fu}{violetfy@umich.edu}
\icmlcorrespondingauthor{Run Peng}{roihn@umich.edu}
\icmlcorrespondingauthor{Honglak Lee}{honglak@eecs.umich.edu \& honglak@lgresearch.ai}

\icmlkeywords{Machine Learning, ICML}

\vskip 0.3in
]



\printAffiliationsAndNotice{}

% \begin{abstract}
% This document provides a basic paper template and submission guidelines.
% Abstracts must be a single paragraph, ideally between 4--6 sentences long.
% Gross violations will trigger corrections at the camera-ready phase.
% \end{abstract}

\begin{abstract}
 Efficient exploration is a challenging topic in reinforcement learning, especially for sparse reward tasks. 
 To deal with the reward sparsity, people commonly apply intrinsic rewards to motivate agents to explore the state space efficiently. 
 In this paper, we introduce a new intrinsic reward design called \Ours{} - \OursName{}, which combines the traditional lifelong novelty motivation with an episodic intrinsic reward that is designed to maximize the stepwise reachability expansion. 
 More specifically, we apply learned world models to generate predicted future states with random actions. 
 States with more unique predictions that are not in episodic memory are assigned high intrinsic rewards. 
 Our method greatly outperforms previous state-of-the-art methods on 12 of the most challenging Minigrid navigation tasks and improves the sample efficiency on locomotion tasks from DeepMind Control Suite. 
\end{abstract}

 % random actions to estimate the volume change of the reachable space with the action and try to maximize this volume. 
% 

\section{Introduction}
Efficient exploration in state space is a fundamental challenge in reinforcement learning (RL)~\cite{hazan2019provably,lee2019efficient}, especially when the environment rewards are sparse~\cite{mnih2013playing,mnih2016asynchronous,schulman2017proximal} or absent~\cite{liu2021behavior,parisi2021interesting}.
Such reward sparsity makes RL algorithms easy to fail due to the lack of useful signals for policy update~\cite{riedmiller2018learning,florensa2018automatic,sekar2020planning}. 
A common approach for exploration is to introduce self-motivated intrinsic rewards such as state visitation counts~\citep{strehl2008analysis, kolter2009near} and prediction errors~\citep{stadie2015incentivizing, pathak2017curiosity, burda2018exploration}.
Most of these intrinsic reward designs measure lifelong state novelty and prioritize visiting states that are less visited starting from the beginning of training. 

% \begin{figure*}[ht]
% \begin{center}
% \centerline{\includegraphics[width=.95\textwidth]{Figures/episodic_buffer_v9_purple.pdf}}
% \vspace{-5pt}
% \caption{An illustration of how our episodic buffer updates. We consider reachable states that are $k=2$ time steps away from $s_t$. After the agent moves from $s_1$ to $s_2$, the episodic buffer $\mathcal{M}$ (shaded in green) expands by 3 new reachable states (shaded in yellow). More formally, $\mathcal{M}  \gets  \mathcal{M} \cup \mathcal{M}_R(s_2)$, where $\mathcal{M}_R(s_2)$ indicates a set of all the states 2-step reachable from $s_2$. Notice that all the states in the trajectory, i.e., $s_0, s_1, s_2$ are also added to the buffer.} 
% \label{fig:method}
% \end{center}
% \vskip -0.3in
% \end{figure*}

\begin{figure*}[ht]
\centering
% \begin{center}
% \includegraphics[width=0.98\textwidth]{Figures/model_description_v5.pdf}
\centerline{\includegraphics[width=0.95\textwidth]{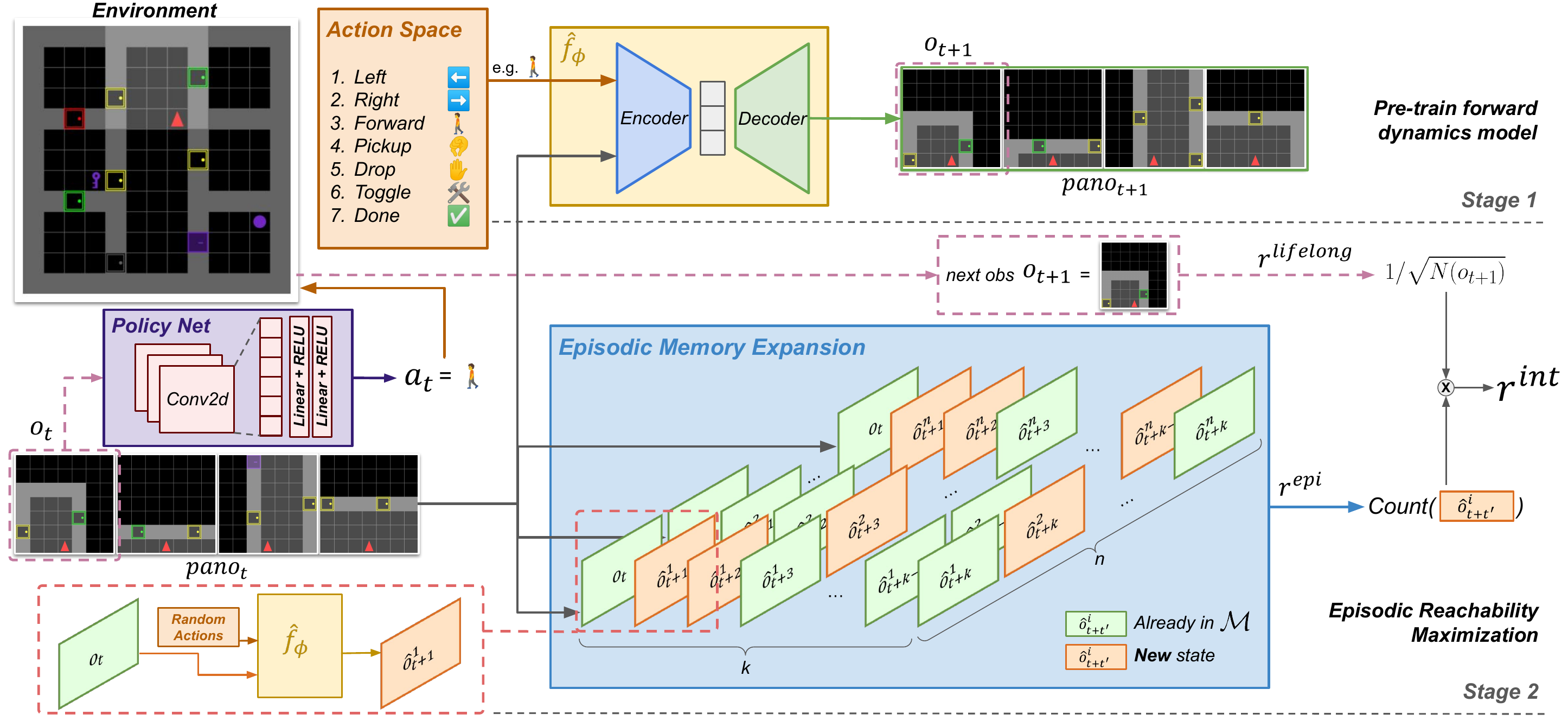}}
\caption{Illustration of how \Ours{} works on Minigrid. For the environment in the upper-left corner, the red triangle indicates the position and orientation of the agent. It has a $7\times7$ partially-observable view (highlighted). 
During pre-training (Stage 1), we collect data using a random policy to train a forward dynamics model $\hat{f}_{\phi}(pano_{t}, a_t) = o_{t+1}$, where $pano_{t}$ denotes the panoramic view as is defined in Section~\ref{para:dynamics_training}. 
For policy training (Stage 2),
we apply $\hat{f}$ to predict observations in the future $k$ time steps with $n$ random actions for each step.
We add the new ones to an episodic buffer $\mathcal{M}$ and take the change of size of $\mathcal{M}$ as the episodic intrinsic reward $r^{\mathrm{epi}}$. The lifelong intrinsic reward is COUNT-based. 
Our intrinsic reward \Ours{} is $r^{\mathrm{lifelong}} * r^{\mathrm{epi}}$. }
% \vspace{-2pt}
\label{fig:model_description}
% \end{center}
\vspace{-5pt}
\end{figure*}

%% HL: Related to generalization of RL. Possibly motivate from that perspective as well.. 
While the above methods achieves great improvement on hard-exploration tasks like Montezuma’s Revenge~\cite{burda2018exploration}, they generally only work well on ``singleton'' environments, where training and evaluation environments are the same. 
However, due to the poor generalization performance of reinforcement learning in unseen environments~\cite{kirk2021survey}, nowadays researchers have been paying more attention on procedurally-generated environments~\citep{cobbe2019quantifying,cobbe2020leveraging,flet2021adversarially}, where the nature of task remains the same but the environment is randomly constructed for each new episode. For example, a maze-like environment will have different maze structures, making it rare for the agent to encounter the same observations across different episodes. 
Therefore, lifelong novelty intrinsic motivations usually fail in hard procedurally-generated environments of this kind~\cite{raileanu2020ride,zha2021rank} because an agent will be trapped around newly-generated states.

% HL: make it clear that in procedurally generated environments, the environment's structure (e.g., maze structure, etc.) and the osbservations from it can be different for different episodes.. 
% blundell2016model,
Inspired by human's frequent use of short-term memory~\cite{andersen2006hippocampus,eichenbaum2017role} to avoid repeatedly visiting the same space, recent work propose to derive intrinsic rewards on episodic level~\cite{savinov2018episodic,badia2020never,raileanu2020ride,zha2021rank,zhang2021noveld}. 
The episodic intrinsic rewards generally give bonus to large episodic-level state space visitation coverage, therefore encourage visiting as many states as possible in the same episode. 
However, \textit{does visiting more states necessarily mean efficient episodic-level exploration}?
We notice that some state visitations are unnecessary and can be avoided if they are predictable from episodic memory. 
For example, when navigating through a house to find a fridge, if you open a door and find an empty room, you do not need to go into it anymore because you can easily predict what the states are like in the room (i.e., intuitively speaking, you would be moving around in an empty room). 
With this inspiration, we propose to design the episodic intrinsic reward to not only maximize the number of visited states in an episode, but also consider those states that are not visited but can be predicted from episodic memory. 

More precisely, 
we maintain an episodic buffer to store all the visited states as well as states reachable from the visited states within a few time steps. 
To get the reachable states, we train a world model with forward dynamics function and apply random actions to the learned dynamics model to predict future states. The predictions are added to the episodic buffer if they are not there already. 
We use the change of size of this episodic buffer as the episodic intrinsic reward. 
Following many previous work, we weight the episodic intrinsic reward by a lifelong intrinsic reward~\cite{badia2020never,zhang2021noveld} like the COUNT-based rewards. 
With this newly proposed intrinsic reward design \textbf{GoBI} - \textbf{Go} \textbf{B}eyond \textbf{I}magination, the agent is expected to both explore the most of the state space throughout training to discover extrinsic rewards, and learn to act in an efficient manner within a single episode to avoid being trapped by seemingly novel states. 

The contributions of this work can be highlighted as follows: 
(i) We propose a novel way to combine world models with episodic memory to formulate an effective episodic intrinsic reward design. 
(ii) In sparse-reward procedurally-generated Minigrid environments~\cite{MinigridMiniworld23}, \Ours{} greatly improves the training sample efficiency in comparison with prior state-of-the-art intrinsic reward functions. 
(iii) \Ours{} extends well to DeepMind Control Suite~\cite{tunyasuvunakool2020} with high-dimensional visual inputs and shows promising results on sparse-reward continuous control tasks. 
(iv) We analyze the design of \Ours{} and present extensive ablations to show the contribution of each component.

\section{Method}
We consider reinforcement learning problems framed as Markov Decision Process (MDP) 
$M = (\mathcal{S}, \mathcal{A}, T, R, \gamma)$, 
where $\mathcal{S}$ and $\mathcal{A}$ denote the state space and action space. 
$T: \mathcal{S}\times \mathcal{A} \times \mathcal{S} \rightarrow [0, 1]$ is the state transition function. 
% $O: \mathcal{S}\times \mathcal{A} \times \Omega \rightarrow [0, 1]$ is the observation function. 
$R: \mathcal{S}\times \mathcal{A} \times \mathcal{S} \rightarrow \mathbb{R}$ is the reward function.
$\gamma$ is the reward discount factor. 
At each step $t$, the state of the environment is denoted as $s_t \in \mathcal{S}$. The agent generates an action $a_t \in \mathcal{A}$ to interact with the environment. The environment then transits to the next underlying state $s_{t+1} \in \mathcal{S}$. 
Apart from the new state $s_{t+1}$, the environment also returns an extrinsic reward $r^{\mathrm{ext}}$ that describes how well the agent reacts to $s_t$. 
In sparse-reward tasks,  $r^{\mathrm{ext}}$ is usually 0.
In this work, we follow the previous work to train RL algorithms with $r^{\mathrm{ext}} + \lambda *
r_t^{\mathrm{int}}$, where $r_t^{\mathrm{int}}$ is a self-motivated intrinsic reward and $\lambda$ is a hyper-parameter that controls the relative importance between intrinsic and extrinsic rewards.

\begin{figure*}[ht]
\begin{center}
\centerline{\includegraphics[width=.95\textwidth]{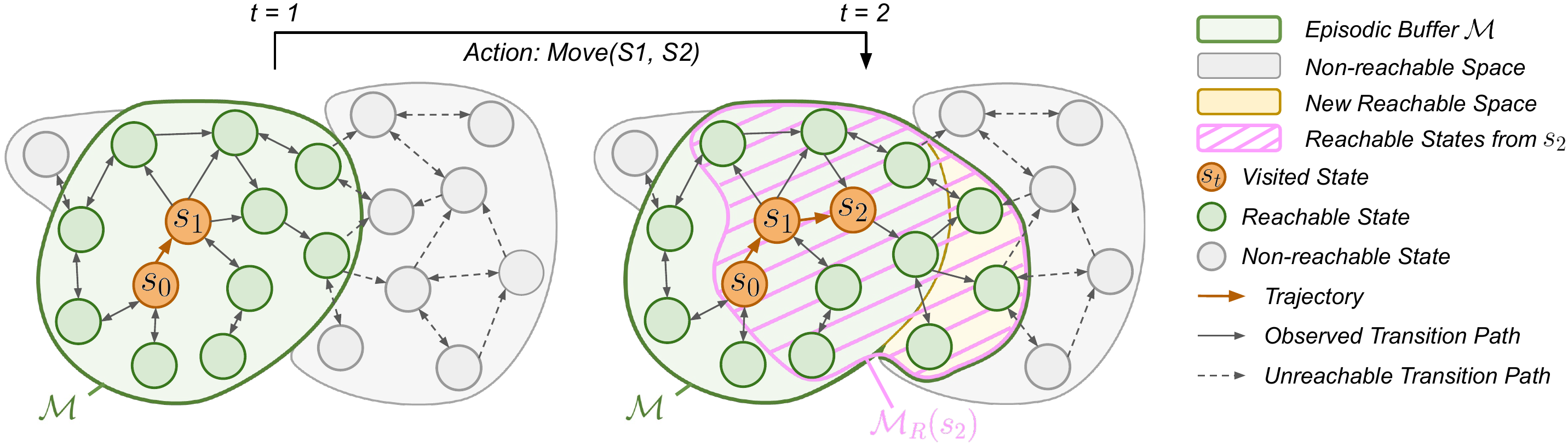}}
\vspace{-5pt}
\caption{An illustration of how our episodic buffer updates. We consider reachable states that are $k=2$ time steps away from $s_t$. After the agent moves from $s_1$ to $s_2$, the episodic buffer $\mathcal{M}$ (shaded in green) expands by 3 new reachable states (shaded in yellow). More formally, $\mathcal{M}  \gets  \mathcal{M} \cup \mathcal{M}_R(s_2)$, where $\mathcal{M}_R(s_2)$ indicates a set of all the states 2-step reachable from $s_2$. Notice that all the states in the trajectory, i.e., $s_0, s_1, s_2$ are also added to the buffer.} 
\label{fig:method}
\end{center}
\vskip -0.3in
\end{figure*}
% HL: might be worth showing k=2 case.. also mention that 
%TODO: change capital S to lowercase s, change unpredicted space to non-reachable space, unpredicted to unreachable, change k to 2, make it more complicated, more nodes/complex structures, show some shaded region to represent k=2}

\subsection{\OursName}
\paragraph{Reachable States and Episodic Buffer}
Our intrinsic reward design aims to exploit the information hidden inside the neighbourhood of states. 
We define a state $A$ to be $k$-step \textit{reachable} from state $B$ if the agent can reach $A$ from  $B$ within $k$ time steps. 
During the training process, for each new episode, we initialize an empty episodic memory buffer $\mathcal{M}$.
At time step $t$, 
we hash $s_t$ as well as all the states reachable from $s_t$.
We denote the set containing all the hash codes of $s_t$ and its reachable states as $\mathcal{M}_R(s_t)$.
Then we update $\mathcal{M}$ by 
$\mathcal{M} \gets \mathcal{M} \cup \mathcal{M}_R(s_t)$. 
Storing the hash codes instead of directly storing the states may alleviate the potential memory issue of the buffer. 
We illustrate this process in Figure~\ref{fig:method}.
When the agent reaches state $s_2$, we add 3 more states that are reachable from $s_2$ but not in $\mathcal{M}$.

\paragraph{Forward Dynamics}
In real environments, it is common that we do not have access to the neighbourhood relationship between states. 
However, we can learn a world model by training a forward dynamics model $\hat{f}_{\phi}(s_t, a_t) = s_{t+1}$ to predict the states reachable from $s_t$. 
This forward dynamics can be pre-trained using data collected by a random policy or trained online together with policy training. 
When training the policy, for each time step $t$, we generate $k\cdot n$ random actions and use the learned dynamics $\hat{f}_{\phi}$ to predict states in the future $k$ steps. 
We hash the current state $s_t$ as well as the predicted future states $\hat{s}^1_{t+1}, ..., \hat{s}^n_{t+1}, ..., \hat{s}^1_{t+k}, ..., \hat{s}^n_{t+k}$ and add the hash codes to the episodic buffer $\mathcal{M}$ if they are not in the buffer. 
Apart from alleviating potential memory issue as is mentioned in the last paragraph, using a hash function may also mitigate the noise introduced by $\hat{f}$. 
With a learned dynamics model, the predictions of reachable states are usually not perfect. However, in the experiment section we show that even with imperfect predictions, our method can improve the training sample efficiency a lot.

\paragraph{Episodic Novelty}
We aim to design an episodic-level novelty reward that guides the agent to extend the frontier of its predicted reachable space efficiently to discover states not visited and not predictable within the same episode. 
More specifically, we denote the size of the episodic buffer $\mathcal{M}$ as $m_t$ at time step $t$ and design a reachability-based bonus $r^{epi} = m_{t+1} - m_t$ that encourages the agent to find unexplored regions. For each time step, the agent is expected to reach the state that is reachable to more new states in the current episode. 

\paragraph{Intrinsic Reward Formulation}
We further weight our episodic intrinsic reward by a lifelong intrinsic reward to encourage the agent to explore the regions that are not well explored in the past.
More formally, the proposed intrinsic reward \Ours{} is defined as:
\vskip -0.2in
\begin{equation}
    r^{\mathrm{int}}_t = (m_{t+1} - m_t) * r_t^{\mathrm{lifelong}} 
    \label{eq:venti}
\end{equation}
% \vskip -0.1in
%
Here, $r_t^{\mathrm{lifelong}}$ denotes lifelong intrinsic reward. We note that our framework is compatible with any choice of lifelong intrinsic reward. Specifically, we use the simple COUNT-based reward $\smash{1/\sqrt{N(s_{t+1})}}$ for the navigation experiments on Minigrid environments \cite{MinigridMiniworld23}, where $N$ denotes the count of $s_{t+1}$ from the start of training.\footnote{For environments that are partially observable (e.g., in Minigrid, the agent observes a 7$\times$7 pixel local view of the environment), we substitute state $s_t$ with observation $o_t$ when calculating the intrinsic rewards.}
For the experiments on DeepMind Control Suite~\cite{tunyasuvunakool2020} we use the state-of-the-art intrinsic reward RE3~\cite{pmlr-v139-seo21a}, which estimates state entropy by a random encoder.

\begin{algorithm}[t]
   \caption{\OursName}
   \label{alg:ours}
\begin{algorithmic}
   \STATE {\bfseries Input:} Intrinsic Reward Coefficient $\lambda_0$, Forward Prediction Step $k$, Number of Random Actions $n$, Intrinsic Reward Decay Parameter $\rho$
   \STATE Initialize policy $\pi_{\theta}$, dynamics model $\hat{f}_{\phi}$, replay buffer $\mathcal{B}$.
   \STATE (Optional) Collect episodes with $\pi_{\theta}$ and train $\hat{f}_{\phi}$ with prediction loss 
   \FOR{episode $e=1,2,...$ until convergence}
   \STATE Initialize episodic buffer $\mathcal{M}$.
   \STATE $\lambda \gets \lambda_0 * ( 1 - \rho) ^ {(e-1)*T}$
   \FOR{$t=1$ {\bfseries to} $T$}
   \STATE Execute $\pi_{\theta}$ in the environment to get a transition pair $(s_t , a_t , s_{t+1} , r^{\mathrm{ext}}_t )$.
   \STATE $m_t \gets size(\mathcal{M})$
   \STATE $\mathcal{M} \gets \mathcal{M} \cup  \{hash(s_t)\}$)
   
   \FOR{$t'=1$ {\bfseries to} $k$}
   
   % \FOR{$i=1$ {\bfseries to} $n$}
    \STATE generate $n$ random actions $a^1_{t'}, ..., a^n_{t'}$
        \STATE $\mathcal{M} \gets \mathcal{M} \cup  \{hash(\hat{f}_{\phi}(\widehat s_{t+t'}^i, a^i_{t'}) | i=1,...,n\}$)
   % \ENDFOR
   \ENDFOR
   % \STATE $\lambda_t \gets \lambda_0 * ( 1 - \rho) ^ {t*e}$
   \STATE $r^{\mathrm{int}}_t = (size(\mathcal{M})-m_t)* r^{\mathrm{lifelong}}$
   \STATE $\mathcal{B} \gets \mathcal{B} \cup \{(s_t , a_t , s_{t+1} , r^{\mathrm{ext}}_t + \lambda * r^{\mathrm{int}}_t)\}$
   \ENDFOR
   \STATE update $\pi_{\theta}$ with RL objective 
   \STATE update $\hat{f}_{\phi}$ with prediction loss 
   
   \ENDFOR
\end{algorithmic}
\end{algorithm}

\paragraph{Intrinsic Decay}
Intrinsic rewards are expected to be asymptotically consistent so that it will not influence the policy learning at later stage of training and result in a sub-optimal policy. 
To guarantee that the policy learning focuses more on extrinsic rewards as training proceeds, in RE3~\cite{pmlr-v139-seo21a}, the authors apply exponential decay schedule for the intrinsic rewards to decrease over time.
Although COUNT-based reward theoretically converges to 0 with enough exploration, it decreases quite slowly in procedurally-generated environments. 
Therefore, we also apply intrinsic reward decay when calculating \Ours{} by decreasing the intrinsic reward coefficient $\lambda$ during training. 
We summarize our method in Algorithm \ref{alg:ours} 
and illustrate the training process on Minigrid navigation tasks in Figure~\ref{fig:model_description}.

\subsection{Conceptual Advantage of \Ours{} over Prior Works}
\begin{figure*}
    \centering
    \includegraphics[width=0.75\textwidth]{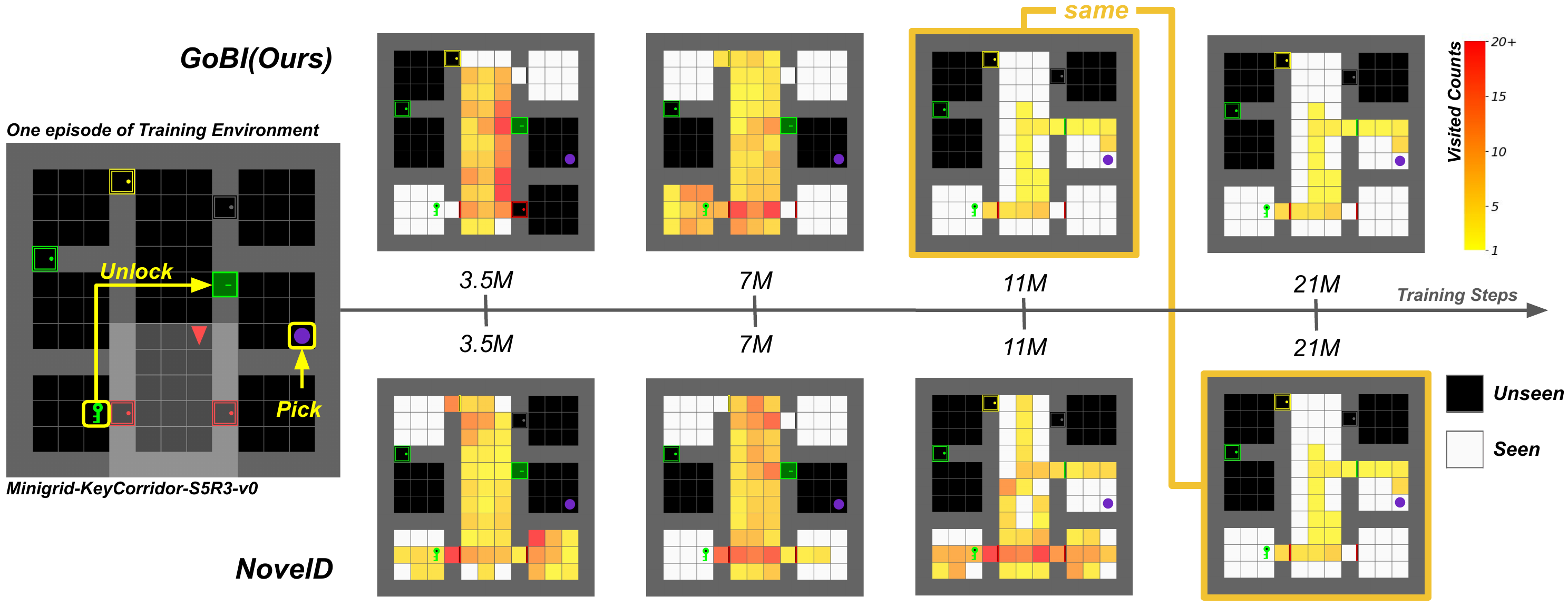}
    \caption{Visitation heatmaps on KeyCorridorS5R3 at different training stages. This figure compares the policy behaviour of \Ours{} and NovelD. A dark red color means plentiful visitations, white means the agent has seen the space but did not step on it, and black means space that are not discovered. 
    It is worth noticing that early in the training (3.5M and 7M time steps), our policy already learns not to go into an empty room, likely because states in an empty room are easily predictable.
    On the contrary, even after 11M steps, an agent trained with NovelD still goes into an empty room (bottom-right corner) for more state visitations.}
    \label{fig:qualitative}
\end{figure*}

Previous works including RIDE~\cite{raileanu2020ride} and NovelD~\cite{zhang2021noveld} also combine episodic intrinsic reward with lifelong novelty as we do. 
However, most of them focus on episodic-level state visitation. For example, NovelD only assigns non-zero rewards to a state when it is visited for the first time in the episode. 
However, we notice that not all state visitations are necessary. The agent's goal for exploration is to gather information about the states. Therefore for states that are easily predictable from episodic memory, visiting them may not really help to acquire more information about the environment. In Figure~\ref{fig:qualitative}, we plot the visitation heatmap of \Ours{} and NovelD to demonstrate the different exploration behaviours of the two methods. 

Our method is closely related to another work that measures episodic curiosity (EC)~\cite{savinov2018episodic}. 
In EC, the authors train a reachability network that takes in two arbitrary states and outputs a similarity score between 0 and 1, where 1 indicates the two states are the same and 0 indicates they are totally different. 
The network is trained using collected episodes by marking temporally close states as positive examples and temporally far ones as negative samples. 
Meanwhile, they also maintain an episodic buffer. 
A state $s_t$ is compared with all the states in the buffer and gets a high intrinsic reward if the corresponding similarity scores are low. 
Only with low enough similarity scores do they add $s_t$ to the buffer. 
Although their method and ours are similar at high level, they are different by design. 
For example, for an agent standing in front of an empty blind alley with dead end, agent trained with \Ours{} does not benefit in going deep into the blind alley because everything there can be predicted as reachable and added to the episodic buffer already. 
However, EC encourages going to the very end of the blind alley to reach the state with low similarity score and high intrinsic reward, even though going into an empty blind alley is not beneficial for exploration and wastes time that can be used to explore other parts of the environment. 
In Appendix~\ref{subsection:ec_heatmap}, we present the visitation heatmaps of policies learned by EC and find that it prefers going to the room corners, which well matches our explanation above. 

\begin{figure}[h]
    \begin{subfigure}{0.49\linewidth}
        \includegraphics[width=\linewidth]{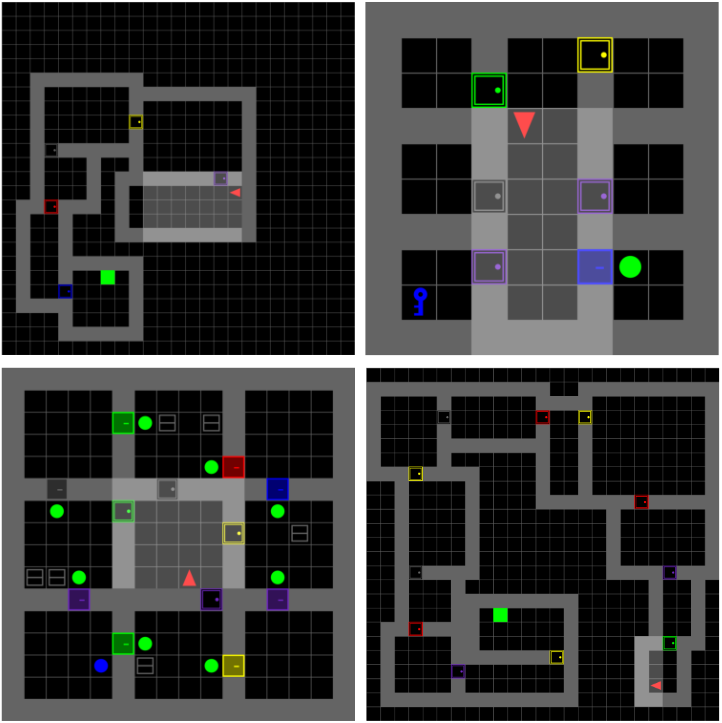}
        \vspace*{-0.15in}
        \caption{MiniGrid}
        \label{fig:fig_minigrid}
    \end{subfigure}
    \hfill
    \begin{subfigure}{0.49\linewidth}
        \includegraphics[width=\linewidth]{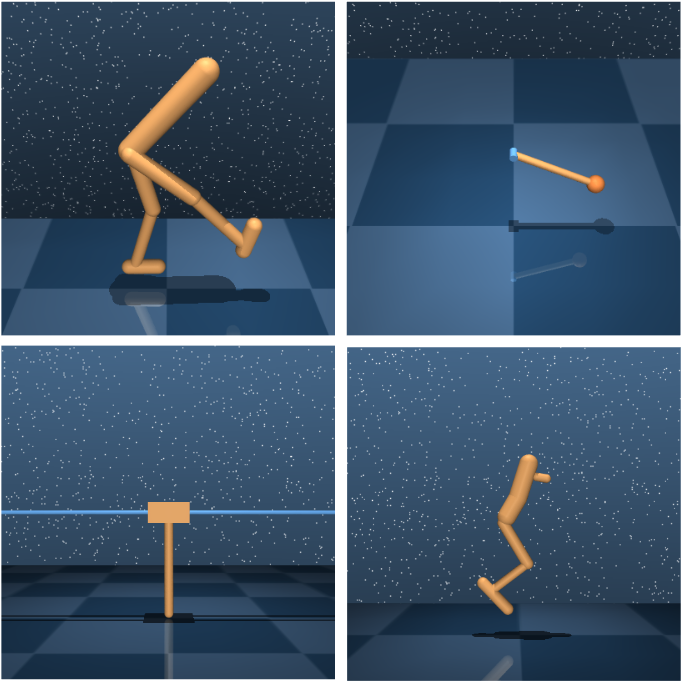}
        \vspace*{-0.15in}
        \caption{Deepmind Control}
        \label{fig:fig_dmcontrol}
    \end{subfigure}
    \vspace*{-0.1in}
    \caption{Rendering of the environments used in this work. Left: 2D grid world navigation tasks that require object interactions. Right: DeepMind Control tasks with visual observations.}
    \label{fig:fig_environment_example}
\end{figure}

\section{Experiments}
\label{sec:experiment}

In this section, we evaluate \Ours{} in two domains: 2D procedurally-generated Minigrid environments~\citep{gym_minigrid} with hard-exploration tasks and locomotion tasks from DeepMind Control Suite~\citep{tunyasuvunakool2020}.
The experiments are designed to answer the following research questions: 
(1) How does \Ours{} perform against previous state-of-the-art intrinsic reward designs in terms of training-time sample efficiency on challenging procedurally-generated environments? 
(2) Can \Ours{} successfully extend to complex continuous domains with high-dimensional observations, for example control tasks with visual observations?
(3) How does each component of our intrinsic reward contribute to the performance?
(4) What is the influence of the accuracy of the learned world models to our method?

\subsection{Minigrid Navigation Tasks}

\paragraph{Minigrid Environments}
MiniGrid \citep{gym_minigrid} is a set of partially-observable procedurally-generated grid world navigation tasks. 
The agent is expected to interact with objects such as keys, balls, doors, and boxes to navigate through rooms and find the goal that is randomly placed in one of the rooms. 
The tasks only provide one sparse reward at the end of each episode, which indicates if the agent successfully finds the goal or not and how many steps it takes to reach the goal. 
In this work, we consider 3 types of tasks including MultiRoom, KeyCorridor, and ObstructedMaze. Some environments that we experiment on in this paper are shown in Figure \ref{fig:fig_minigrid}. 
The upper-right is a KeyCorridor-S4R3 environment, where the agent should learn to open the doors to find a key, use it to open the locked blue door, and pick up the green ball. 
The bottom-left figure shows an ObstructedMaze-Full environment, which is similar to KeyCorridor but more challenging. The rooms are larger, the doors are blocked by balls, and the keys are hidden in boxes. 
The upper-left and bottom-right environments are MultiRoom environments, in which the agent has to navigate through connected rooms to reach the goal in the last room.

\paragraph{Baselines}
We compare with state-of-the-art intrinsic reward designs that work well on Minigrid including NovelD~\cite{zhang2021noveld}, RIDE~\citep{raileanu2020ride}, and RND~\citep{burda2018exploration}.
For a fair comparison, we follow the same basic RL algorithm and network architectures used in the official codebase of NovelD and only change the intrinsic rewards $r^{\mathrm{int}}$ for all the methods. 
We also compare our method with EC~\cite{savinov2018episodic} because of the similarity of the high-level idea between the two methods. 
However, the original paper of EC does not include experiments on Minigrid. Therefore, we implement our own version to adapt to Minigrid. 
We follow their implementation suggestions in the paper and tune the hyper-parameters such as novelty threshold by grid search. 

\begin{figure*}
    \centering
    \includegraphics[width=0.8\textwidth]{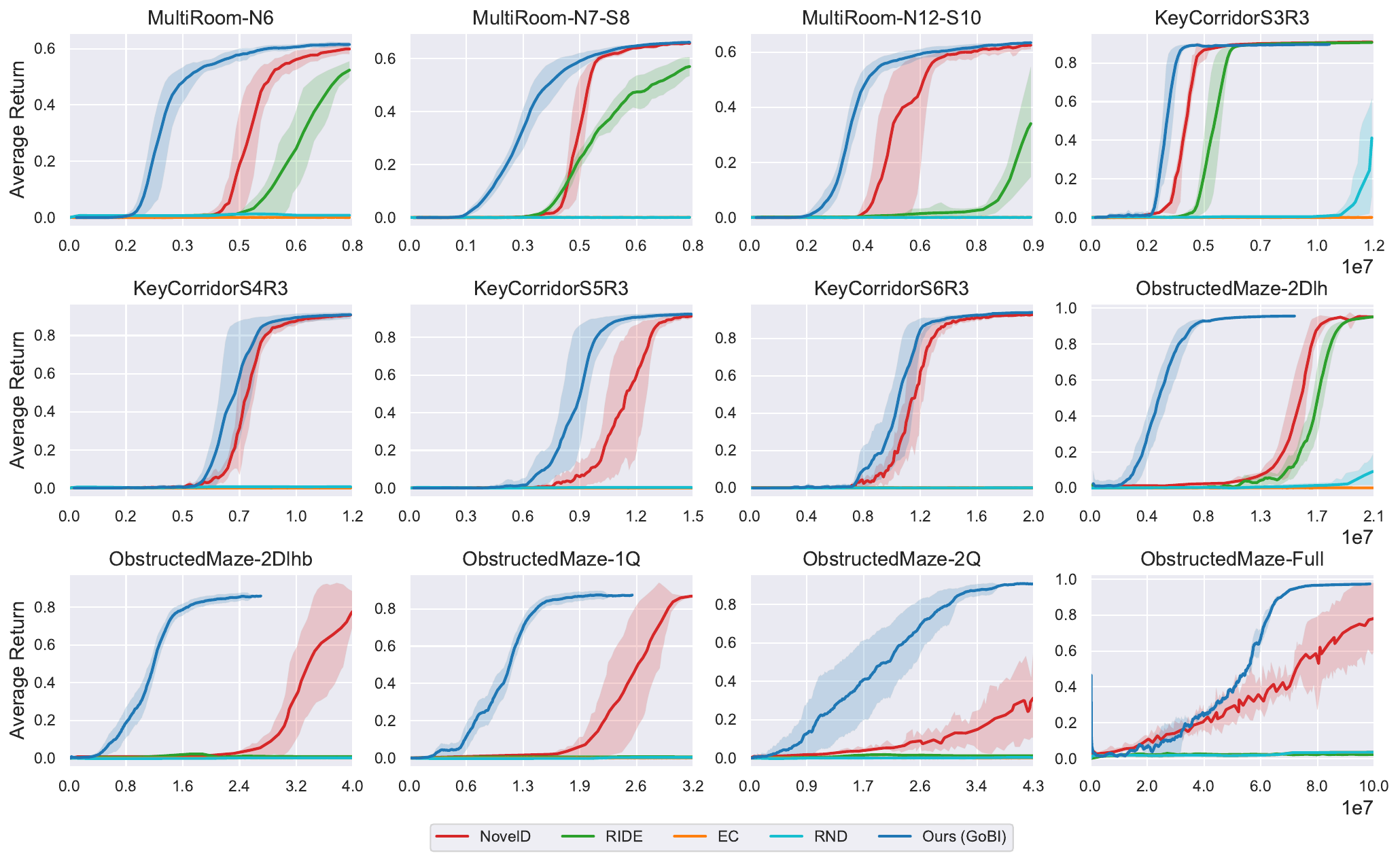}
    \caption{Training performance of \Ours{} and the baselines on 12 MiniGrid environments. The x-axis shows the number of environment steps. We shift the training curves towards right by the number of environment steps we use to pre-train the dynamics model, i.e. $1\mathrm{e}5$ time steps. Results are averaged across 4 seeds. }
    \label{fig:minigrid_train}
\end{figure*}

\paragraph{Dynamics Model Training} \label{para:dynamics_training}
For each experiment on Minigrid, we first run a random policy for $1\mathrm{e}5$ steps to collect data and use them to train a forward dynamics model as the world model. Among the pairs collected, there are about $5\mathrm{e}4$ different transition pairs. 
%During the experiments we notice that whether fine-tuning the pre-trained dynamics model or not during policy training does not influence the performance much.
During our experiments, we observe that fine-tuning the pre-trained dynamics model during policy training has no significant influence on the performance.
Similar to \cite{parisi2021interesting}, we use the $360^{\circ}$ panoramic views as the input to predict the future observations. This is a rotation-invariant representation of the observed state. We consider this still a fair comparison with the previous state-of-the-arts because both NovelD and RIDE rely on using the state information instead of observations for the episodic count calculation. 

Due to the limited field of view of the agent, we only forward the learned dynamics by $k=1$ step when predicting. We predict the next observations produced by all 7 discrete actions in the Minigrid tasks including turn left, turn right, forward, toggle, pick up, drop, and done. 
We directly apply the default Python hashing function to hash the observations and predicted future observations. We do not expect the hashing function to mitigate the prediction error on Minigrid, but only use it to reduce the dimension of observations and predictions.

\paragraph{Training Performance on Minigrid}
Figure \ref{fig:minigrid_train} shows the learning curves of \Ours{} and state-of-the-art exploration baselines NovelD, RIDE, RND, and EC on 12 most challenging Minigrid navigation tasks, including MultiRoom, KeyCorridor, and ObstructedMaze. Our curves are shifted towards right by the number of random exploration environment steps used to train the world model. 
In all 12 environments, \Ours{} significantly outperforms previous methods in terms of sample efficiency. For instance, on ObstructedMaze-2Dlhb, \Ours{} is about three times more sample efficient than NovelD. On the hardest ObstructedMaze-Full environment, \Ours{} achieves near-optimal performance within 70M steps. 
Lastly, although we try to tune the hyper-parameters of EC, our implementation of EC still does not learn well on the Minigrid environments.

\paragraph{Qualitative Results}
To clearly present the exploration behavior learned by \Ours{}, we show the visitation heatmaps of \Ours{} and NovelD on a KeyCorridorS5R3 environment in Figure~\ref{fig:qualitative}. Not only does our method converge to an optimal policy faster, the exploration behaviour is very different from NovelD. \Ours{} quickly learns not to visit easily predictable states like an empty room, making it more efficient to explore interesting parts of the environment, for example, the room with a key in it.

\subsection{Experiments on Control Tasks}
We further test \Ours{} on DeepMind Control Suite, which are a set of image-based continuous control tasks. 
These tasks are more challenging than Minigrid because of its high-dimensional observations and stochastic transitions. 
Notice that these environments are not procedurally-generated. The experiments in this section are to show the generality of our method by experimentally showing that \Ours{} extends well to sparse-reward tasks with continuous action space and high-dimensional observation space. 
% The experiments in MuJoCo allow us to show the generality of our method.

\paragraph{Dynamics Model Training} 
We follow the world model structure in Dreamer \cite{hafner2019dream} and directly apply their encoder, transition model, and observation model to predict future observations. 
However, compared to Minigrid, it requires way more data to train a decent dynamics model on DeepMind Control to generate visually-reasonable predictions. 
Therefore, different from the experiments on Minigrid, we do not pre-train the dynamics models. 
Instead we train the dynamics model together with the policy as is shown in Algorithm \ref{alg:ours}. 
We find that the number of sampled random actions $n=5$ works well across all 4 environments. 
For the number of forward prediction steps $k$, we set it to be 3 for Pendulum Swingup and 1 for the other 3 environments.
% Note that $n$ and $k$ are set as hyper-parameters, but we find that in most of our experiments $k=1$ is sufficient for efficient exploration. 
% % HL: is the performance robust/sensitive to hyperparameters? maybe show in the appendix? -> will do that 

\begin{figure}[ht]
% \vskip 0.2in
\begin{center}
\centerline{\includegraphics[width=\columnwidth]{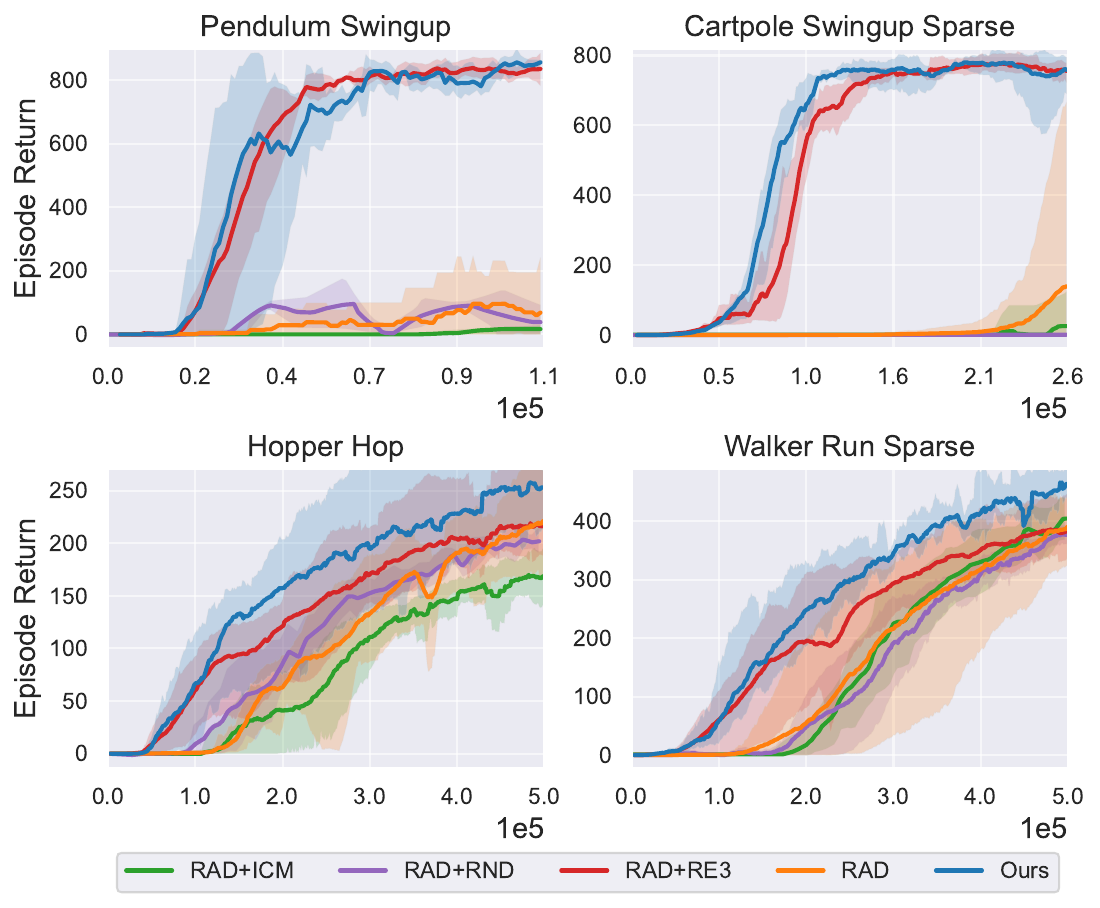}}
\caption{Training curves of \Ours{} and the baselines on DeepMind Control Suite. The curves are averaged across 5 seeds. }
\label{fig:dmc_train}
\end{center}
\vskip -0.3in
\end{figure}

For the hashing function, we find that a simple SimHash as is suggested in \cite{tang2017exploration} works well in capturing the similarities between similar observations. We use SimHash to hash the 
% $84\times84\times3$ 
image observations to 50 bits.

\paragraph{Training Performance on DeepMind Control}
We compare with the state-of-the-art intrinsic motivation on DeepmMind Control tasks - RE3 \cite{pmlr-v139-seo21a}, which applies a k-nearest neighbor entropy estimator in the low-dimensional representation space of a randomly initialized encoder to maximize state entropy. RE3 is also what we use for the lifelong intrinsic reward part $r^{\mathrm{lifelong}}$ of \Ours{} in Eq~\ref{eq:venti}.
Another two intrinsic reward baselines we consider are ICM~\cite{pathak2017curiosity} and RND~\cite{burda2018exploration}. 
For a fair comparison, all the experiments use the same basic RL algorithm RAD~\cite{laskin2020reinforcement}. 
The results are shown in Figure \ref{fig:dmc_train}. The additional episodic-level intrinsic reward term improves the sample efficiency a lot compared to only using lifelong intrinsic reward, especially on Hopper Hop and Walker Run Sparse.

\subsection{Ablation Study}
\paragraph{\Ours{} Variations}
\label{subsection:ablation}
In this section, we analyze how each component of our intrinsic reward contributes to the final performance. 
We ablate each component of \Ours{} and run experiments on Minigrid environments with the following:

\begin{itemize}
\vspace*{-0.1in}
\setlength\itemsep{-0.2em}
% \item \Ours{}: $\smash{(m_{t+1} - m_t)/\sqrt{N(o_{t+1})}}$
\item R1: only episodic intrinsic reward $m_{t+1} - m_t$
\item R2: indicator of whether new states are added to the episodic buffer
$\smash{(\mathbbm{1}\{m_{t+1}-m_t>0\})/\sqrt{N(o_{t+1})}}$
\item R3: only lifelong intrinsic reward $\smash{1/\sqrt{N(o_{t+1})}}$
% \vspace*{-0.05in}
\end{itemize}

\begin{figure}[ht]
\begin{center}
\centerline{\includegraphics[width=\columnwidth]{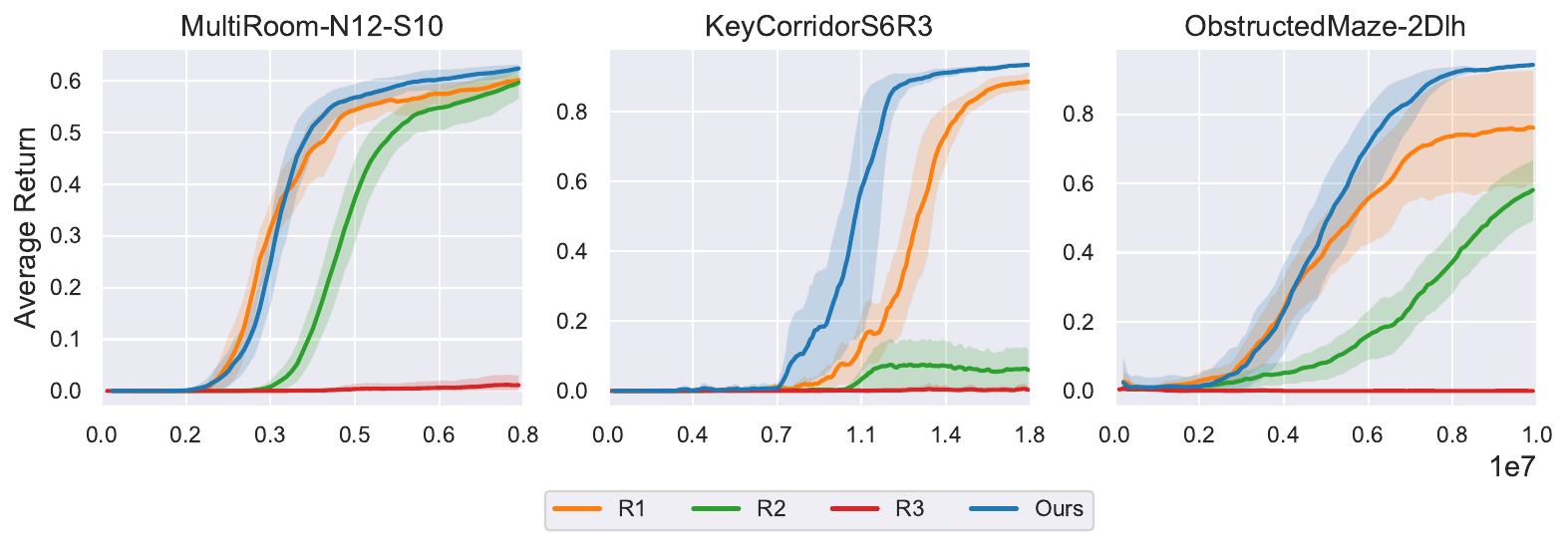}}
\caption{Training performance comparison among \Ours{}, R1, R2, and R3 on 3 Minigrid environments. 
}
\label{fig:ablation}
\end{center}
\vskip -0.2in
\end{figure}

Training performance of \Ours{} as well as R1, R2, and R3 are shown in Figure~\ref{fig:ablation}. Although R1 works on MultiRoom, it suffers on Obstructed Maze and large KeyCorridor environments. 
The underlying reason may be that in Key Corridor and Obstructed Maze the room structures change less across episodes than MultiRoom (all generated rooms are squares with fixed sizes), therefore the COUNT-based rewards contribute more in such environments than in MultiRoom. 
At the same time, R2 performs way worse than \Ours{}. 
Agents trained with R2 prefer actions that only increase the size of the episodic buffer a bit therefore getting positive score more often. We provide an illustrative example in Appendix~\ref{subsection:ablation_illus} to explain why R2 does not work well compared to \Ours{}. 
Using only lifelong intrinsic reward R3 performs the worst and struggles to learn efficiently on large Multiroom, Key Corridor, and Obstructed Maze environments. 
% The reason behinds observation (1) is that in Obstructed Maze environments, there are a lot more possible states (larger state space) therefore the agent will learn to achieve very high reachability score and focus less on reaching the real goal. It is hard to balance the intrinsic and external reward in this case. 

\begin{figure}[ht]
\begin{center}
\centerline{\includegraphics[width=0.8\columnwidth]{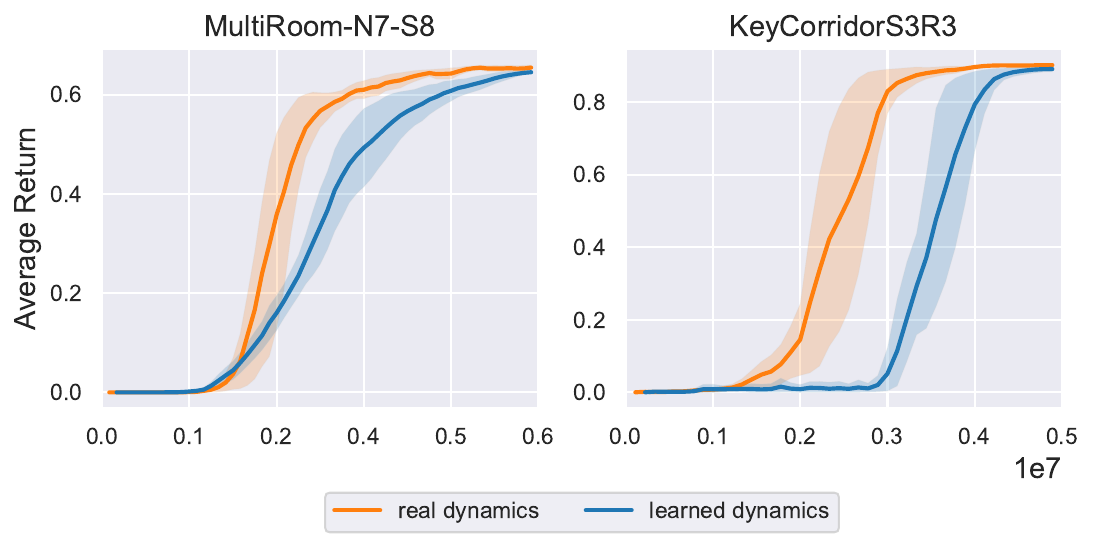}}
\caption{Comparison between using the real dynamics model of the environments vs using a learned one on Minigrid environments. In both MultiRoom and KeyCorridor, using a real dynamics model to derive intrinsic reward makes the policy converge faster, especially on KeyCorridor. }
\label{fig:forwardx}
\end{center}
\vskip -0.3in
\end{figure}

\paragraph{Real Dynamics vs Learned Dynamics}
A learned dynamics model is generally not perfect, especially for partially-observable environments like Minigrid. 
In many cases the predictions can never be accurate. 
For example, when the agent first opens the door of a new room, usually it will not accurately predict everything behind the door. 
Figure~\ref{fig:forwardx} shows the training curves between using the real dynamics model vs a learned dynamics model. 
Not surprisingly, with the same intrinsic reward function, using the real dynamics converges faster to a near-optimal policy. 
However, even with imperfect dynamics model, our method still greatly surpasses previous state-of-the-arts. 

\begin{figure}[ht]
\begin{center}
\centerline{\includegraphics[width=0.8\columnwidth]{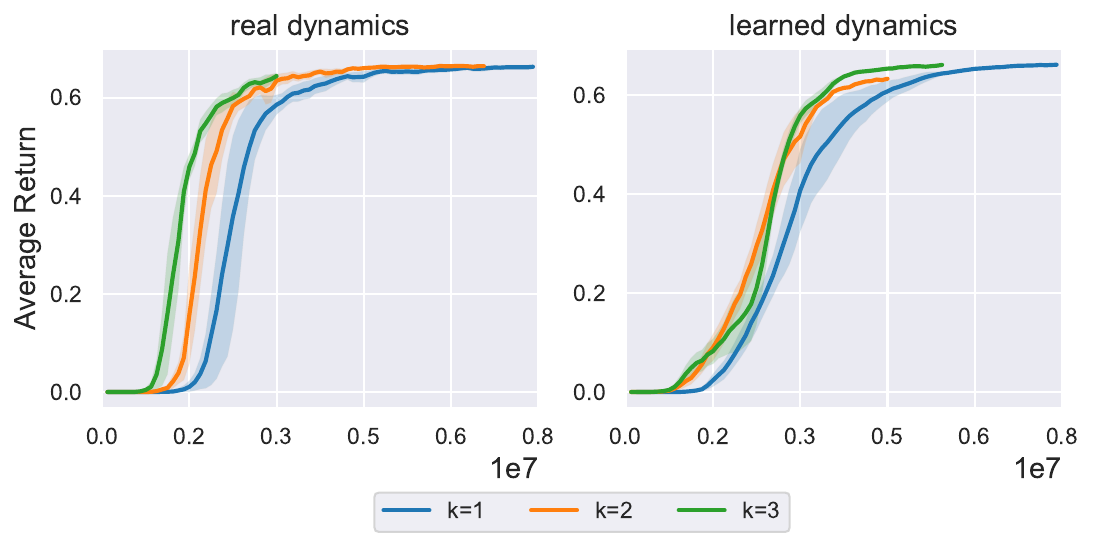}}
\caption{We make forward predictions for different number of future steps $k$ using both the real dynamics and the learned dynamics model. The plots above show the training performance on Minigrid MultiRoom-N7-S8. }
\label{fig:k_hyper}
\end{center}
\vskip -0.3in
\end{figure}

\paragraph{Multi-Step Predictions}
Figure ~\ref{fig:k_hyper} shows the learning performance of \Ours{} on Minigrid with a varying choices of the number of future steps to do predictions $k=1,2,3$. 
For $k>1$, our dynamics model outputs $pano_{t+1}$ instead of $o_{t+1}$ and we hash and store the observations from panoramas in each future time step. 
%When $k=0$, our $r^{epi}$ is identical to the episodic visitation in NovelD (but counting observation $o_t$ instead of state $s_t$). 
%The performance drops significantly compared to non-zero choices of $k$. 
With a real forward dynamics model, a larger $k$ generally accelerates exploration more, because it prioritizes actions that lead to the states that are reachable to more states in the long run. 
However, due to the limited field of view of the agent and the model inaccuracy, this is not the case if we use a learned model. Forwarding 2 steps is still faster than only 1 step, but more steps than that does not really make exploration faster.

% \vskip -0.5in
\section{Related Work}
\subsection{Exploration in Reinforcement Learning}
\vspace*{-0.08in}
Efficient exploration in reinforcement learning, especially for sparse-reward reinforcement learning problems is challenging. 
A natural and popular solution is to design some metric to evaluate state novelty and assign high intrinsic reward to novel states. 
% self-motivated intrinsic rewards to guide exploration. 
For example, COUNT-based intrinsic reward \citep{strehl2008analysis, kolter2009near, tang2017exploration} and curiosity-based intrinsic motivation \cite{stadie2015incentivizing, pathak2017curiosity, burda2018exploration}. 
Another popular way is to do state space entropy maximization \cite{hazan2019provably, lee2019efficient}. 
Recently, nearest neighbor entropy estimation methods \cite{yarats2021reinforcement, liu2021behavior} have shown great performance improvements in challenging visual domains. 
Our method is compatible with all these successful exploration intrinsic reward designs by using them as $r^{\mathrm{lifelong}}$, but we additionally encourage the episodic-level reachable space expansion to achieve large state space coverage within a single episode. 
% Some prior works aim to maximize information gain for exploration \citep{storck1995reinforcement, little2013learning, mobin2014information, shyam2019model}. 

\subsection{Episodic Memory}
\vspace*{-0.08in}
Deriving useful information from episodic buffer have shown great success in improving the training sample efficiency in RL on navigation, control, and Atari games.  
Episodic memory buffers are applied to mimic hippocampal episodic control and rapidly assimilate recent experience \cite{blundell2016model, pritzel2017neural}. 
As is mentioned in the previous sections, \cite{savinov2018episodic} keeps an episodic buffer to store observations and introduce an episodic curiosity module to determine if a new observation is reachable from previous observations or not. 
RAPID \cite{zha2021rank} proposes a novel way to do behaviour cloning on episodes with high episodic coverage. 
NGU \cite{badia2020never} combines an episodic novelty module and a lifelong novelty module to generate intrinsic rewards. 
However, in NGU, the episodic novelty is a measurement of difference between the current observations from the previous observations, while ours focus on how much the reachable space is expanded from the new state. 
RIDE~\cite{raileanu2020ride} and NovelD~\cite{zhang2021noveld} both count the episodic state visitations, while we claim that apart from visited states, we should also consider states that can be predicted from short-term episodic memory. 

%\subsection{Learning World Models}
\subsection{Learning World Models with Forward Dynamics}
\vspace*{-0.08in}
%\subsection{Learning World Models via Forward Dynamics}
%\subsection{Forward Dynamics Training}

Learning 
% world models with 
dynamics function from 
a set of observed data
% $\mathcal{D} = \{(s_t, a_t, s_{t+1})\}$ 
is a widely-studied topic in reinforcement learning, especially due to the rapid growth of model-based reinforcement learning \cite{wang2019benchmarking}. 
Existing work show that an agent's world model is implicitly a forward model that predict future states~\cite{ha2018recurrent, freeman2019learning}. 
% Learned world models enable agent to imagine future states with a sequence of actions~\citep{NEURIPS2018_2de5d166}. 
% For tasks with partial observability, one of the most popular solutions is to use recurrent neural networks \cite{chiappa2017recurrent, gemici2017generative, ha2018recurrent}. 
Recently, people have proposed latent dynamics models that work well on high-dimensional inputs~\cite{okada2020planet}.
These latent dynamics models encode image observations and predict future states in the latent space~\cite{ha2018world, hafner2019dream, hafner2023mastering}, outputting realistic future observations on visually complex domains
including DeepMind Control Suite \cite{tunyasuvunakool2020}, VizDoom \cite{kempka2016vizdoom}, Atari Games, and DeepMind Lab \cite{beattie2016deepmind}. 
The learned dynamics models can be used to guide exploration by prediction error \cite{stadie2015incentivizing, pathak2017curiosity, burda2018exploration}, surprise \cite{achiam2017surprise}, or information gain by variance of model ensemble means \cite{sekar2020planning}. 
Our method differ from the previous methods by directly generating and hashing the predicted states and add them to an episodic reachable state buffer. With the advanced world model structures, our method can be extended to diverse domains with complex observations. 

% Dreamer\cite{laskin2020reinforcement} then introduce a more powerful structure to imagine future states in the latent space and do long-horizon planning. These structures have successful performance on complex control and Atari games, which make our idea of calculating the forward reachability space volume possible.  
% \cite{hafner2019dream}

%\vspace*{-0.08in}
\section{Discussions and Future Work}
% \vspace*{-0.05in}
This paper shows an effective way to combine learned world models with episodic memory to intrinsically guide efficient exploration.
% s the agent to both find the states that are reachable to more unexplored states in the episode and to avoid visit easy-to-predict states that does not provide useful information of the environment. 
% We demonstrate that by combining the agent's knowledge of the environment dynamics with its episodic memory
Our method achieves state-of-the-art performance on procedurally-generated hard exploration tasks and also works well on singleton continuous control domains.
However, it still has certain limitations. 
First of all, the dynamics model we use for the Minigrid experiments is deterministic, making it possible to generate less accurate predictions and making the performance of our method worse than using the real dynamics. 
A possible way to make improvement on this is to make the prediction model generative and sample possible future states. 
Secondly, for the control tasks with complex visual inputs, we hash the images with static hashing to make them discrete hash codes. However, to better capture the semantic similarities between the image observations, it would be beneficial to learn hash functions, for example, by using an autoencoder (AE) to learn meaningful hash codes~\cite{tang2017exploration}. We leave these investigations as future work. 
\section{Conclusion}
\vspace*{-0.05in}

In this work, we introduce \OursName - \Ours, a novel episodic intrinsic reward design that encourages efficient episodic-level exploration by expanding reachable space. While most previous episodic intrinsic rewards use a naive episodic state count or state visitation coverage, our method exploits learned world models to predict reachable states and motivates the agent to seek for the states with more unexplored neighbors. Combined with lifelong intrinsic rewards, our method shows great training time sample efficiency improvement on hard procedurally-generated environments. At the same time, it can be extended to guide exploration on continuous control tasks with visual inputs, both indicating a promising future in this direction.

\section{Acknowledgments}
This work was supported in part by grants from LG AI Research, NSF IIS 1453651, and NSF FW-HTF-R 2128623.


\bibliography{reference}
\bibliographystyle{icml2023}

%%%%%%%%%%%%%%%%%%%%%%%%%%%%%%%%%%%%%%%%%%%%%%%%%%%%%%%%%%%%%%%%%%%%%%%%%%%%%%%
%%%%%%%%%%%%%%%%%%%%%%%%%%%%%%%%%%%%%%%%%%%%%%%%%%%%%%%%%%%%%%%%%%%%%%%%%%%%%%%
% APPENDIX
%%%%%%%%%%%%%%%%%%%%%%%%%%%%%%%%%%%%%%%%%%%%%%%%%%%%%%%%%%%%%%%%%%%%%%%%%%%%%%%
%%%%%%%%%%%%%%%%%%%%%%%%%%%%%%%%%%%%%%%%%%%%%%%%%%%%%%%%%%%%%%%%%%%%%%%%%%%%%%%
% \newpage
% \appendix
% \onecolumn

\newpage
\clearpage
\appendix
\section{Implementation Details}
\label{app:imp_detail}
%e.g. data split, hyper-parameters, required in checklist
%e.g. the total amount of compute and the type of resources used (e.g., type of GPUs, internal cluster, or cloud provider), required in checklist
\subsection{Experiments on Minigrid} 
\label{app:implementation-minigrid}

\paragraph{Baselines}
Implementations of \Ours{}, NovelD~\cite{zhang2021noveld}, RIDE~\cite{raileanu2020ride}, RND~\cite{burda2018exploration}, and EC~\cite{savinov2018episodic} are built on the official codebase of NovelD.  
For fair comparisons, only the intrinsic reward $r^{\mathrm{int}}$ differs among the methods and they all use the same base algorithm IMPALA~\cite{espeholt2018impala}. At the same time, all the experiments are run with the same compute resource with Nvidia TITAN X GPU and 40 CPUs.
For NovelD, we rerun their official code to get the results of Minigrid MultiRoom and KeyCorridor. 
For the experiments on ObstructedMaze, we did not find the proper hyper-parameters to fully reproduce their results. 
Therefore, we directly take the results reported in their paper. 
For RIDE and RND, we run the code in the official codebase of NovelD. 
For EC~\cite{savinov2018episodic}, their original paper does not include experiments on Minigrid environments. Therefore, we implement our own version and tune the hyper-parameters with grid search. 
% A recent paper called MADE~\cite{zhang2021made} also report state-of-the-art performance on Minigrid. However, due to the lack of support for Minigrid experiments in their official codebase and becuase we did not manage to reproduce their results following their implementation details, we do not include them in the experiment section. 
% Our results on Minigrid are still better than what they report in the the paper though. 
The intrinsic reward functions of \Ours{} and the baselines are listed below:
\begin{itemize}
    \item \Ours{}: $(m_{t+1}-m_t)/\sqrt{N(o_{t+1})}$, where $m_t$ is the size of the episodic buffer $\mathcal{M}$ and $N(o_{t+1})$ is the lifelong count of the observation $o_{t+1}$ starting from the beginning of training. 
    \item RND: $\|\phi(o_{t+1})-\hat{\phi}(o_{t+1})\|^2$, which is the difference between a fixed random network $\hat{\phi}$ and a trained state embedding network $\phi$. Here, $\phi$ is trained to minimize the same error.
    \item NovelD: $\max[novelty(o_{t+1})-\alpha \cdot novelty(o_t), 0]*\mathbbm{1}\{N^{epi}(s_{t+1}=1)\}$.
    They apply RND to measure the novelty of $o_t$, i.e., $novelty(o_t) = \|\phi(o_t)-\hat{\phi}(o_t)\|^2$. $N^{epi}(s_{t+1}=1)$ checks if the agent visits state $s_{t+1}$ for the first time in an episode. Notice that they use the full environment information, i.e. everything in the grid world instead of only the $7\times7$ partially-observable view. Therefore $N^{epi}$ counts $s_{t+1}$ instead of $o_{t+1}$.
    \item RIDE: $\|\phi(o_t)-\phi(o_{t+1})\|^2/\sqrt{N^{epi}(s_{t+1})}$, where $\phi$ is the state embedding network trained to minimize the prediction error of an inverse and a forward dynamics. $N^{epi}$ indicates the episodic counts. Same as NovelD, in RIDE, they also use the state information $s_{t+1}$ for episodic count. 

    \item EC: $\beta - {C(\mathcal{M}, o_{t+1}})$, where $C(\mathcal{M}, o_{t+1})$ is the 90-th percentile similarity scores between $o_{t+1}$ and all the observations in the episodic buffer $\mathcal{M}$. The similarity scores are calculated using a pre-trained episodic curiosity module. $\beta$ is a hyper-parameter.
\end{itemize}

% \vspace*{-0.06in}
\paragraph{Policy and Value Function Training}
For fair comparisons, the policy network and value function network are the same for all approaches. 
The input observations of dimension $7\times 7\times 3$ are put into a shared feature extraction network, which includes three convolutional layers of kernel size$=3\times 3$, padding$=1$, channel=$32,128,512$, and stride$=1,2,2$ respectively with ELU activation.
The features are then flattened and put through 2 linear layers with 1024 units and ReLU activation, and an LSTM layer with 1024 units. 
This shared feature is passed separately to 2 fully-connected layers with 1024 units to output action distribution and value estimation.

\paragraph{Dynamics model} 
For our implementation of the dynamics model, our input is the panorama of the current step. 
To get the panorama, we let the agent rotate for 3 times and concatenate the 4 observations to get inputs of size $28 \times 7 \times 3$. It is then passed to a feature extraction module that has the same structure as our policy and value function networks, except that the input to the first linear layer is $4\times1024$. 
We then concatenate it with actions and put it through a decoder with 2 linear layers of sizes 256 and 512, and reshape back to $7\times7\times3$ to get a predicted observation. We pre-train the dynamics model using $1\mathrm{e}{5}$ $(pano_t, a_t, o_{t+1})$ pairs collected by a random policy. 
RIDE also requires training dynamics models for the state embedding network $\phi$. 
The input of their dynamics model is the state embedding and action. 
The forward model contains two fully-connected layers with 256 and 128 units activated by ReLU. The inverse dynamics model contains two fully-connected layers with 256 units and a ReLU activation function. Its input is the state embeddings of two consecutive steps. 

\paragraph{Hash Functions}
We directly apply the default Python hashing function to hash the $7\times7\times3$ observations and predicted future observations before adding them to the episodic buffer.

% \vspace*{-0.06in}
\paragraph{State embedding} NovelD, RIDE, and RND all require training a state embedding network $\phi$. 
The input is the observation in MiniGrid with dimension $7\times 7\times 3$.
It contains three convolutional layers with kernel size$=3\times 3$, padding $=1$, stride $=1,2,2$, number of channels $=32, 128, 512$ respectively. The activation function is ELU. Following the convolutional layers are two linear layers of 2048 and 1024 units with ReLU activation. 

% \vspace*{-0.06in}

\vspace*{-0.06in}
\paragraph{Visitation Count} 
For \Ours, $N(o)$ stores the flattened $7\times 7 \times 3$ observations of each step. And for NovelD and RIDE, they count the full states at episodic level, whose shape varies from environment to environment. For example, the shape is $25\times 25 \times 3$ for MultiRoom environments.  

% \vspace*{-0.06in}
\paragraph{Hyper-parameters}
Table~\ref{tab:hyp_driven} shows the values of hyper-parameters shared across different methods.
\begin{table}[!ht]
\centering
\begin{tabular}{c|c}
\toprule
\makecell{Parameter name} & \makecell{Value} \\ 
\midrule
Batch Size & 32\\
Optimizer & RMSProp\\
Learning Rate & 0.0001\\
LSTM Steps & 100\\
Discount Factor $\gamma$ & 0.99\\
Weight of Policy Entropy Loss & 0.0005 \\
Weight of Value Function Loss & 0.5 \\
\bottomrule
\end{tabular}
\caption{Hyper-parameters for experiments on Minigrid. These hyper-parameters are shared across all the methods}
\label{tab:hyp_driven}
\end{table}

For all of our experiments using \Ours{} on Minigrid, we set the intrinsic reward coefficient $\lambda=0.01$ and $k=1$, which means only forwarding the dynamics model by $1$ step. 
At the same time, as the action space is small and discrete, instead of randomly sampling some actions, we directly predict the future observations using all $7$ possible actions. 
We list the hyper-parameter choices of intrinsic decay factor $\rho$ in Table~\ref{tab:hyp-minigrid}. The value of $\rho$ is chosen to make the intrinsic reward large at the beginning of training and near-zero at the end of the training. 

\begin{table}[!ht]
\centering
\begin{tabular}{c|c}
\toprule
\makecell{Parameters} & \makecell{Value} \\ 
\midrule
Forward Step $k$ & 1\\
Intrinsic Decay $\rho$ & $6\mathrm{e}{-7}$ for MR-N7S8;\\
& $8\mathrm{e}{-7}$ for MR-N12S10, MR-N6;\\
& $1.5\mathrm{e}{-6}$ for KC-S3R3;\\
& $5\mathrm{e}{-7}$ for KC-S4R3, KC-S5R3\\
& $3\mathrm{e}{-7}$ for KC-S6R3, OM-2Dlh\\
& $2\mathrm{e}{-7}$ for OM-1Q, OM-2Dlhb, OM-2Q\\
& $5\mathrm{e}{-8}$ for OM-Full\\
$\hat{f}_{\phi}$ Optimizer & Adam\\
$\hat{f}_{\phi}$ Learning Rate & $5\mathrm{e}{-4}$\\
\bottomrule
\end{tabular}
\caption{The hyper-parameters of \Ours{} for experiments on Minigrid.}
\label{tab:hyp-minigrid}
% \vspace*{-0.3in}
\end{table}

For NovelD, we set $\lambda=0.05$ for all the environments as is suggested in their official codebase.
% on ObstructedMaze environments and $\alpha=0.1$ on all other environments.
For RIDE, we use $\lambda=0.1$ on KeyCorridor-S3R3 and $\lambda=0.5$ on all other environments.
For RND, we set $\lambda=0.1$ on all the environments. 
For EC, 
% following the authors' suggestions, we choose $\beta$ to be $0.5$ for Minigrid environments with varying episode length. For the intrinsic reward coefficient 
we make $\lambda=0.01$ so that the initial average intrinsic reward of EC is similar to ours. 

% In RND, following \cite{campero2020learning}, we use the intrinsic reward coefficient $\alpha=0.1$ for all the environments.

\subsection{Experiments on Deepmind Control Suites} 
\paragraph{Baselines}

Implementations of \Ours{}, RE3~\cite{pmlr-v139-seo21a}, ICM~\cite{pathak2017curiosity}, and RND~\cite{burda2018exploration} are built on the official codebase of RE3. 
All the experiments apply the same base reinforcement learning algorithm RAD~\cite{laskin2020reinforcement}.
For RE3, we rerun their official code to get the results on all four environments. For ICM and RND, we follow the implementation details listed in RE3 to implement them to be compatible with DeepMind Control tasks. 
For a fair comparison, only the intrinsic reward design differs among the methods. 
The intrinsic reward functions of \Ours{} and the baselines are listed below:
\begin{itemize}
    \item \Ours{}: $(m_{t+1}-m_t) \times \mathrm{log}(||y_i - y_i^{k-NN}||_2 + 1)$, where $m_t$ is the size of the episodic buffer $\mathcal{M}$. The latter part is the RE3 intrinsic reward which we introduce below. 
    \item RE3: $\mathrm{log}(||y_i - y_i^{k-NN}||_2 + 1)$, where $y_i = f_{\theta}(s_i)$ is a fixed representation outputs from a randomly initialized encoder and $y_i^{k-NN}$ is a set of k-nearest neighbors of $y_i$ among all the collected $y$'s from the beginning of training. 
    \item ICM: $\frac{\eta}{2}||\hat{\phi}(o_{t+1})-\phi(o_{t+1})||_2^2$, where $\eta$ is a scaling factor. $\phi(o)$ is a feature vector that is jointly optimized with a forward prediction model and an inverse dynamics model and $\hat{\phi}(o)$  predicts the feature encoding at time step $t+1$.
    % is the predicted estimate of $\phi(o)$.
    \item RND: $\|\phi(o_{t+1})-\hat{\phi}(o_{t+1})\|^2$, which is the difference between a fixed random network $\hat{\phi}$ and a trained state embedding network $\phi$. Here, $\phi$ is trained to minimize $\|\phi(o_{t+1})-\hat{\phi}(o_{t+1})\|^2$.
\end{itemize}

\paragraph{Architecture}
The observation size of all the environments is $84\times84\times3$. The encoder architecture follows the same one as in~\cite{yarats2021improving}, which contains 4 convolutional layers of $3 \times 3$ kernels, channel=$32$, and stride=$2,1,1,1$ with ReLU activations. The output is then passed to a fully-connected layer and normalized by LayerNorm.

\paragraph{Dynamics Model}
For the forward dynamics model that we use to generate future predictions, we apply the same world model structure as in Dreamer~\cite{hafner2019dream}. The input size of Dreamer is $64\times64\times3$. We down-sample the input observations to $64\times64$ instead of tuning the world model layers. 
We train the dynamics model together with the RL policy in an online manner instead of pre-train it because it takes many episodes for the predictions to be visually reasonable. Therefore it does not add extra effort to determine how many data should we collect to pre-train the dynamics model. 

\paragraph{Image hashing}
As the observations are images in high-dimensional space and the predictions are usually not accurate, we hash the images to lower dimension to avoid taking too much space and to collapse similar observations and predictions.
Following~\cite{tang2017exploration}, we use the simple SimHash function to map the images to 50 bits. More specifically, we project the flattened images to a random initialized vector and use the signs of output vector values as the hash code. 

\paragraph{Hyper-parameters}
Table~\ref{tab:dmc-hyper} shows the values of hyper-parameters shared across different methods.
\begin{table}[!h]
\centering
\begin{tabular}{c|c}
\toprule
\makecell{Parameter name} & \makecell{Value} \\ 
\midrule
Augmentation & Crop\\
Observation Size & (84, 84)\\
Action Repeat & 2\\
Replay Buffer Size & 100000\\
Initial Random Exploration Steps & 1000\\
Frame Stack & 3 \\
Actor Learning Rate & 0.0002 \\
Critic Learning Rate & 0.0002 \\
Batch Size & 512 \\
\# Nearest Neighbors & 3 \\
Critic Target Update Freq & 2 \\
\bottomrule
\end{tabular}
\caption{Hyper-parameters for experiments on DeepMind Control Suites. These hyper-parameters are shared across all the methods}
\label{tab:dmc-hyper}
\end{table}

For the intrinsic reward coefficients, we follow the best choices reported in RE3. For \Ours, we apply the same intrinsic reward coefficient $\lambda$ and intrinsic reward decay $\rho$ as the ones in RE3 for fair comparison. The intrinsic rewards that are specific to our method is shown in Table~\ref{tab:dmc-hyper2}. For the number of random actions $n$, we perform hyper-parameter search over \{3, 5, 10, 20\} and find that $n=5$ perform well across all the tasks. For the number of forward steps $k$, we perform hyper-parameter search over \{1, 2, 3, 5\} and report the ones with the best results.  

\begin{table}[!ht]
\centering
\begin{tabular}{c|c}
\toprule
\makecell{Parameter name} & \makecell{Value} \\ 
\midrule
\# Forward Step $k$ & 3 for pendulum-swingup;\\
& 1 for others\\
\# Random Actions $n$ & 5\\

\bottomrule
\end{tabular}
\caption{Hyper-parameters for experiments on DeepMind Control Suites. These hyper-parameters are specific to our method. }
\label{tab:dmc-hyper2}
\end{table}

\section{Hyper-Parameters}
\subsection{Intrinsic reward decay}

\begin{figure}[h]
\begin{center}
\centerline{\includegraphics[width=0.5\columnwidth]{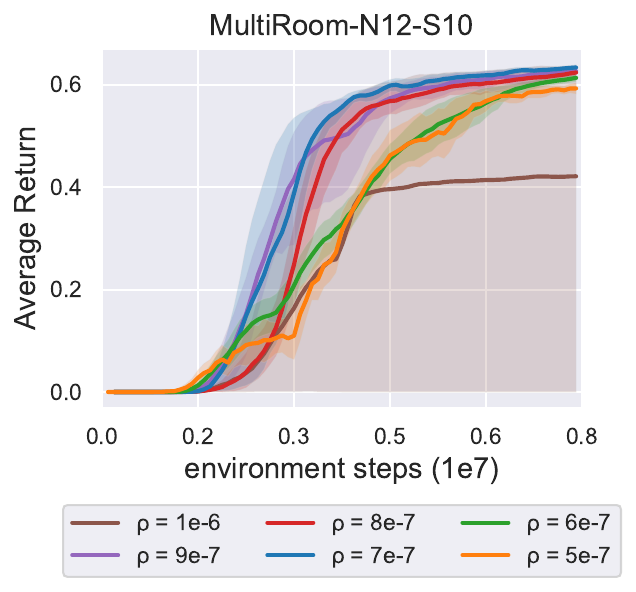}}
% \vspace{-5pt}
\caption{Training performance of \Ours{} on Minigrid Multiroom-N12-S10 with different intrinsic reward decay $\rho$.}
\label{fig:rho_hyper}
\end{center}
% \vskip -0.2in
\end{figure}

In Figure~\ref{fig:rho_hyper}, we show the training performance with different intrinsic reward decay $\rho$. The choice of $\rho$ is to balance between the relative importance of the extrinsic and intrinsic reward. If $\rho$ is too large, for example when $\rho=1\mathrm{e}-6$, before the agent finds any goal, the intrinsic reward already decreases to very small. Therefore sometimes it is hard for the agent to learn anything useful, resulting in unsatisfactory performance. Meanwhile, if $\rho$ is too small, for example when $\rho=5\mathrm{e}-7$, the intrinsic reward will be too large at later stage of training and make the agent focus less on the extrinsic reward. Therefore the policy may converge slower.

% An effective way to tune this hyper-parameter is to also record the number of $(x, y)$ positions that the agent visits. If the visited area is large but the average return is low, we know that the intrinsic reward is too large and the agent ignores the extrinsic reward. 

\begin{figure}[h]
\begin{center}
\centerline{\includegraphics[width=0.5\columnwidth]{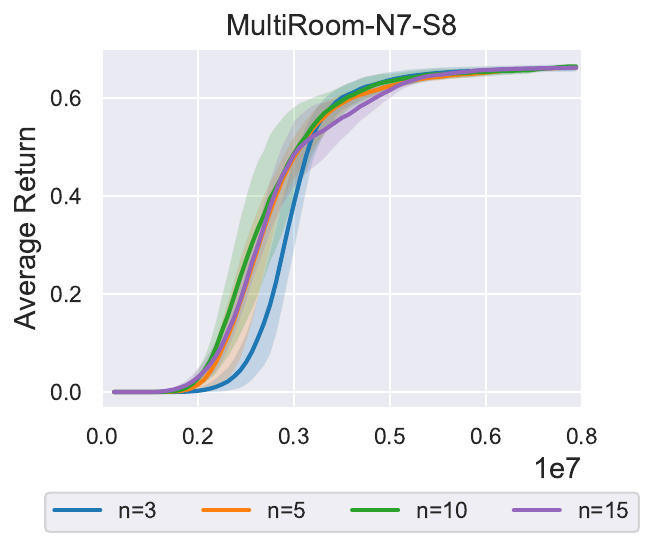}}
\vspace{+5pt}
\caption{Training performance of \Ours{} on Minigrid Multiroom-N7-S8 with different $n$ randomly sampled actions per step.}
\label{fig:nactions}
\end{center}
% \vskip -0.2in
\end{figure}

\begin{figure*}[th]
\centering
\includegraphics[width=.75\textwidth]{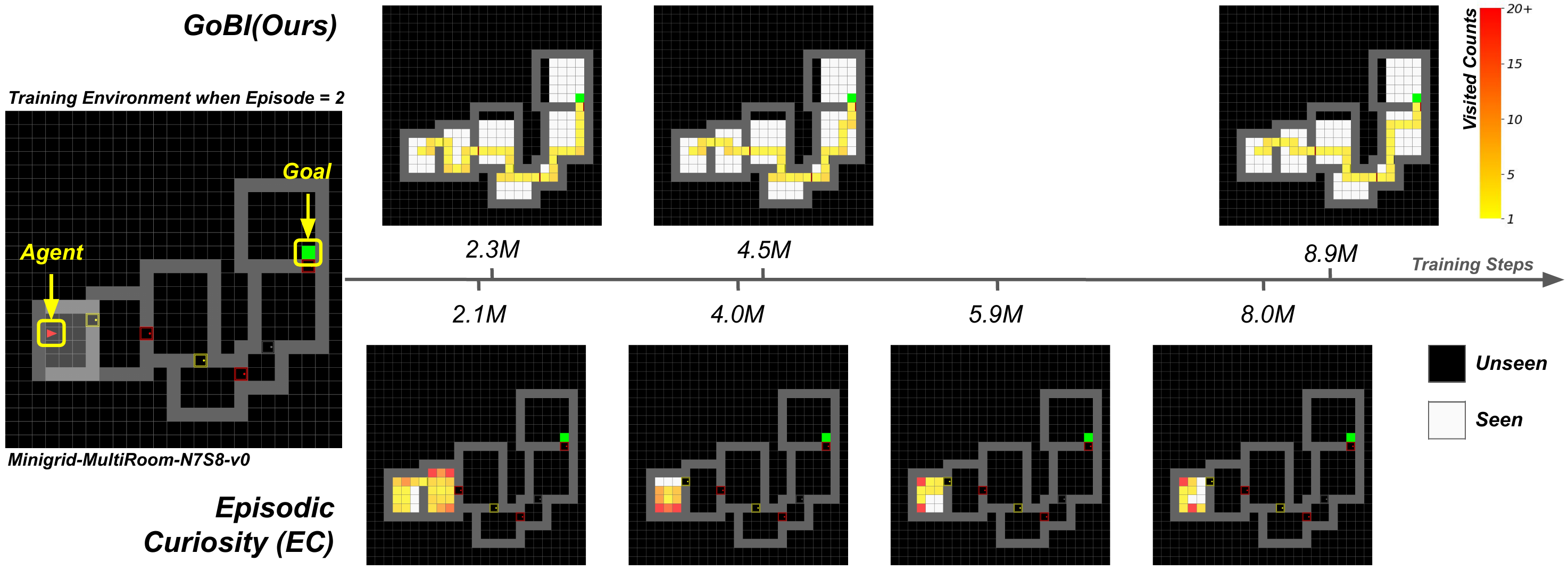}
\caption{Heatmaps of example trajectories at different training steps in Minigrid-Multiroom-N7-S8 environment.}
\label{fig:ec_vs_predx_ep2}
\end{figure*}

\subsection{Number of randomly sampled actions}
In the Minigrid experiments reported in Section~\ref{sec:experiment}, we do not randomly sample actions because Minigrid has a small discrete action space with only 7 actions. Therefore we directly predict future observations of all 7 actions. However, we also report the results with $n=3,5,10,15$ random actions in Figure~\ref{fig:nactions}. To summarize, $n=5,10,15$ all have similar performance on Minigrid, while a larger n makes the wallclock training time longer. 
$n=3$ is slightly slower at the early stage, but still outperforms the previous state-of-the-arts. Overall, $n=5$ would be a good choice for the Minigrid environments.

\section{Exploration Behaviour Comparison between \Ours{} and EC}
\label{subsection:ec_heatmap}

In Figure~\ref{fig:ec_vs_predx_ep2}, we visualize the policy visitation heatmaps of \Ours{} and EC~\cite{savinov2018episodic} on a MultiRoom environment from Minigrid. 
Although EC fails to learn an optimal policy, we can still capture its preference from the heatmaps. 
An agent trained with EC prefers going to the corners of the room, which generally have lower similarity scores than the states in the middle of the room. 
However, if the similarity scores are not low enough for the states to be added to the episodic buffer, it will continue staying at the states to maximize its intrinsic reward. 
We tried to tune the similarity score threshold using grid search but still have not find a good hyper-parameter choice for it because the similarity score at different corners does not share a consistent value. 
Unlike EC, we can see from the figure that \Ours{} chooses not to visit the border of the room early on in training, as the information on the border are easily predictable from the information in the middle of the room.

\begin{figure}[ht]
\begin{center}
\centerline{\includegraphics[width=0.8\columnwidth]{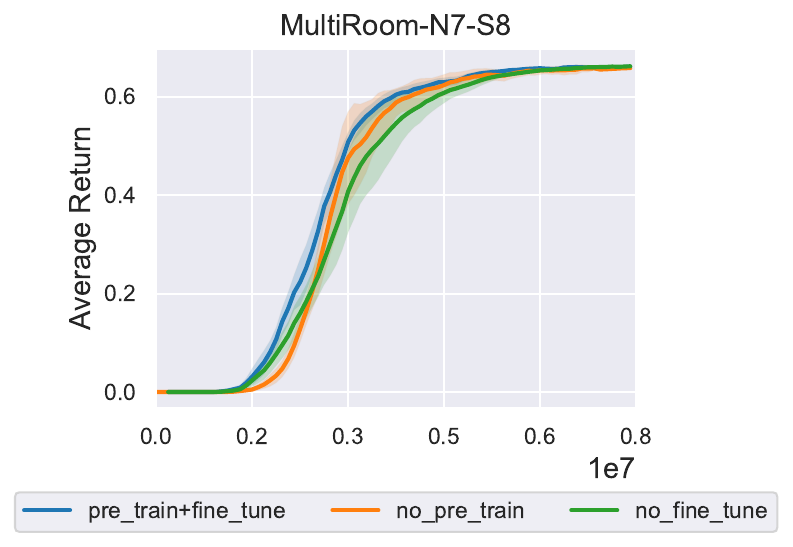}}
\caption{Training performance comparison on MultiRoom-N7-S8 environment among 1) use a fixed pre-trained dynamics model, 2) use a pre-trained dynamics model, and fine-tune it online, 3) no pre-training, directly train the dynamics model together with policy training. }
\label{fig:pretrain_finetune}
\end{center}
\vskip -0.3in
\end{figure}

\section{Wallclock Training Time}
Table~\ref{tab:wallclock} shows the wallclock time needed to train NovelD and our method for 10M Minigrid environment steps. 
GoBI requires about 2x the wallclock time needed to train NovelD for the same number of environment steps. 

\begin{table}[!ht]
\centering
\begin{tabular}{c|c}
\toprule
\makecell{Algorithm} & \makecell{Wallclock Time (hours)} \\ 
\midrule
NovelD & $5.46(\pm0.058)$\\
GoBI(ours) & $10.65(\pm0.082)$\\
\bottomrule
\end{tabular}
\caption{Wallclock training time comparison in hours between NovelD and GoBI. }
\label{tab:wallclock}
\end{table}

\begin{figure}[ht]
\begin{center}
\centerline{\includegraphics[width=\columnwidth]{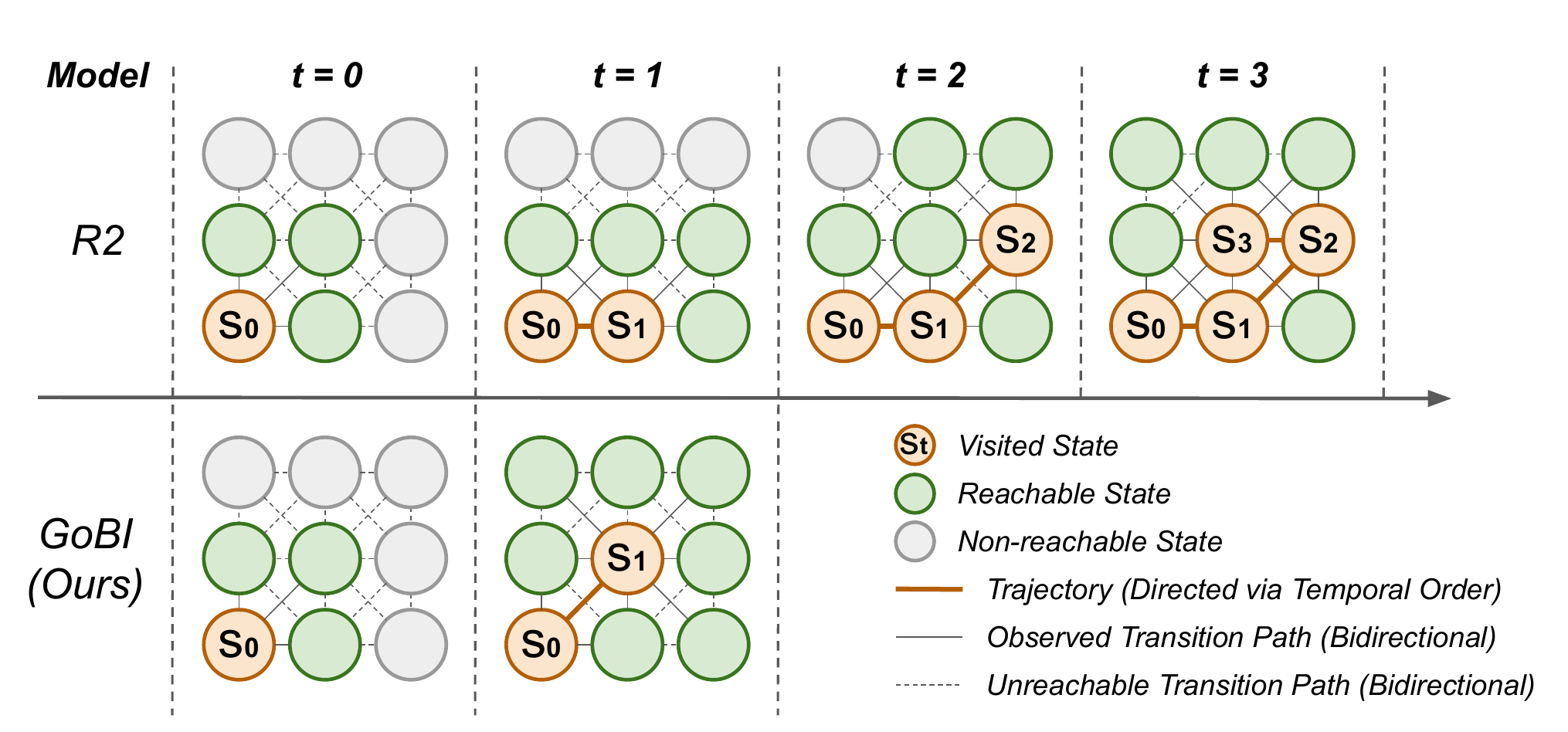}}
% \vspace{-5pt}
\caption{An illustrative example of why R2 does not work well to encourage efficient exploration. In this example, our goal is to include all the states into the episodic buffer quickly. \Ours{} can move directly to the center in one step for maximum intrinsic reward, while R2 may choose to take extra steps for exploration since it mainly focuses on whether the episodic buffer expands or not.}
\label{fig:ablation_illus}
\end{center}
% \vskip -0.2in
\end{figure}

\section{Dynamics Training}
In the experiment section~\ref{sec:experiment}, we report the results of applying a pre-trained forward dynamics for GoBI on Minigrid. However, the forward dynamics model can also be trained together with policy training. In Figure~\ref{fig:pretrain_finetune}, we report the results of an ablation study on a Minigrid environment of 3 settings: 1) use a pre-trained dynamics model, and keep it fixed when training the policy, 2) pre-train a dynamics model, and fine-tune it when training the policy, 3) no pre-training, directly train the dynamics model in an online manner. In summary, all 3 versions work similarly, but due to the fact that training the dynamics model online will add extra wall clock training time, we use option 1 in our main experiments.

\section{Ablation Study Illustration}
\label{subsection:ablation_illus}

In this section, we provide an illustrative example of why only considering whether new states are added to the episodic buffer or not, i.e., R2 in Section~\ref{subsection:ablation}, works way worse than our method. The example is shown in Figure~\ref{fig:ablation_illus}. If these 9 states are only a small part of the environment, we want a policy that explore this part as quickly as possible - mark all the states as reachable as quickly as possible. However, in order to maximize its step-wise intrinsic reward, an agent trained with R2 will go along the border to only add a few new states to the episodic buffer at a time, which wastes many unnecessary steps so is not beneficial for exploration.

% \section{Robustness to Noisy-TV Problem}

%%%%%%%%%%%%%%%%%%%%%%%%%%%%%%%%%%%%%%%%%%%%%%%%%%%%%%%%%%%%%%%%%%%%%%%%%%%%%%%
%%%%%%%%%%%%%%%%%%%%%%%%%%%%%%%%%%%%%%%%%%%%%%%%%%%%%%%%%%%%%%%%%%%%%%%%%%%%%%%

\end{document}